\documentclass{article}

\usepackage{PRIMEarxiv}

\usepackage[utf8]{inputenc} 
\usepackage[T1]{fontenc}    
\usepackage{hyperref}       
\usepackage{url}            
\usepackage{booktabs}       
\usepackage{amsfonts}       
\usepackage{nicefrac}       
\usepackage{microtype}      
\usepackage{lipsum}
\usepackage{fancyhdr}      
\usepackage{graphicx}       
\graphicspath{{media/}}     

\usepackage{amsthm}
\usepackage{amsmath}
\usepackage{amssymb}
\usepackage{verbatim}
\usepackage{mathtools}
\usepackage{float}
\usepackage{bbm}
\usepackage{multirow}
\usepackage{enumitem}   
\usepackage{lscape} 
\usepackage{method}

\DeclareMathOperator*{\argmin}{arg\,min}

\theoremstyle{plain}
\newtheorem{definition}{Definition}

\title{Towards a Shapley Value Graph Framework for Medical peer-influence
\thanks{\textit{\underline{Jamie Duell is supported by the UKRI AIMLAC CDT, funded by
  grant EP/S023992/1.}}: 
} 
}
\author{
  Jamie Duell, Monika Seisenberger, Gert Aarts \\
  Swansea University \\
  \texttt{\{853435,m.seisenberger, g.aarts\}@swansea.ac.uk} \\
   \And
  Shangming Zhou \\
  University of Plymouth\\
  Plymouth\\
  \texttt{shangming.zhou@plymouth.ac.uk} \\
     \And
  Xiuyi Fan \\
  Nanyang Technological University \\
  Singapore\\
  \texttt{xyfan@ntu.edu.sg} \\
}

\begin{document}
\maketitle

\maketitle 

\begin{abstract}
eXplainable Artificial Intelligence (XAI) is a sub-field of Artificial Intelligence (AI) that is at the forefront of AI research. In XAI, feature attribution methods produce explanations in the form of feature importance. People often use feature importance as guidance for intervention.
However, a limitation of existing feature attribution methods is that there is a lack of explanation towards the consequence of intervention. In other words, although contribution towards a certain prediction is highlighted by feature attribution methods, the relation between features and the consequence of intervention is not studied.
The aim of this paper is to introduce a new framework, called a peer influence framework to look deeper into explanations using graph representation for feature-to-feature interactions to improve the interpretability of black-box Machine Learning models and inform intervention. 

\keywords{Explainable AI \and Graph Explanation \and Medical Study \and Feature Attribution \and Interpretability.}
\end{abstract}

\section{Introduction}

Explainable Artificial Intelligence (XAI) is at the forefront of Artificial Intelligence (AI) research with a variety of techniques and libraries coming to fruition in recent years, e.g., model agnostic explanations \cite{Ribeiro2,nori2019interpretml}, counter-factual explanations \cite{Poyiadzi2020FACEFA,ravaramind2}, contrastive explanations \cite{NEURIPS2018_c5ff2543} and argumentation-based explanations \cite{RAGO2021103506,FanXiuyi}. XAI methods are ubiquitous across fields of Machine Learning (ML), where the trust factor associated with applied ML is undermined due to the black-box nature of methods. Generally speaking, a ML model takes a set of inputs (features) and predicts some output; and existing works on XAI predominantly focus on understanding relations between features and output. 

These approaches in XAI are successful in many areas as they suggest how an output of a model might change, should we change its inputs. Thus, {\em interventions} - manipulating inputs in specific ways with the hope of reaching some desired outcome - can be provoked using existing XAI methods when they are capable of providing relatively accurate explanations \cite{Duell,Holzinger+2020+171+179}. However, with existing XAI holding little knowledge to consequences of interventions \cite{Barocas}, such intervention could be susceptible to error. From both a business and ethical stand-point, we must reach beyond understanding relations between features and their outputs; we also need to understand the influence that features have on one another. We believe such knowledge holds the key to deeper understanding of model behaviours and identification of suitable interventions.

In the medical field, where data can be both sensitive and volatile, understanding of peer-influence, which describes the associative impact of one feature on another upon intervention, is important to inform intervention on controllable risk factors \cite{Draper06} to rectify an instance, e.g. a patient outcome. For instance, with existing XAI methods, we may determine that the administered regimen is a significant contributor towards lung cancer survival; at the same time, it also has associative effects on other features, such as weight \cite{Sarcev} or tumour cell proliferation \cite{SenYang}. In the sense that, {\em changing the administered regimen is going to not only affect lung cancer survival, it also changes how ``weight'' and ``tumour cell proliferation'' are  affecting lung cancer survival.}

To fully understand the prediction model, we must be able to reveal relationships between features, both from the core data through extended new or meta-data, and examine how features respond to interventions. In this work, we 
\begin{enumerate}
\item{introduce peer-influence explanations (PI-Explanations) to model feature interactions,} 
\item{introduce peer-influence graphs (PI-Graphs) to visualise PI-Explanations,}
\item{illustrate how effective treatment can be identified from PI-Explanations and PI-Graphs.}
\end{enumerate}
The rest of this paper is organised as follows. 
Section~\ref{proposed} presents our approach. Section~\ref{sec:result} presents two case studies. 
Section~\ref{sec:related} presents a few works on graph-based XAI. 
Section~\ref{sec:cln} gives the conclusion.

\section{Proposed Framework} \label{proposed}

In medicine, we know that understanding the local influence - how specific feature values affecting the prediction made from these values - 
holds greater importance for clinicians \cite{Tonekaboni}. 
Thus, in introducing the peer-influence explanation framework, 
we would like to leverage on existing explainable methods to provide insight and 
utilise the feature attribution values 
computed with XAI methods such as e.g., Local Interpretable Model-Agnostic Explanation (LIME)\cite{Ribeiro2}, Diverse Counterfactual Explanations (DiCE)\cite{nori2019interpretml} or SHapley Additive Predictions (SHAP)\cite{Lundberg}, 
as they provide {\em local explanations} to prediction instances. 

With our approach, we aim to explain how changes to each feature $x_i$ in 
an instance $\mathbf{x}$ can affect other features in $\mathbf{x}$, by utilising 
feature attribution values $\phi$ returned through some XAI method. 
Specifically, for each feature, we study its influence to all other features, by eliminating its impact and seeing how other features' contributions to the prediction change. Also, note that, in this article we use approximated Shapley values calculated from the feature attribution method SHAP for its convenience; 
our approach can conceptually be adapted to any XAI method returning feature attribution values.

The first key element in our work is the notion of {\em Peer-Influence Explanations}. 
To introduce this, we start by defining a few notions, starting with {\em feature explanation} as follows. 

\begin{definition}
  \label{dfn:feature_explanation}
  Given an instance $\mathbf{x} = [x_1, \ldots, x_m]$, let $\phi =
  [\phi_1, \ldots, \phi_m]$ be the feature attribution explanation of $\mathbf{x}$ computed
  with some feature attribution explainer $g$. Then for each $x_i$ in
  $\mathbf{x}$, its {\em feature explanation} (computed using $g$) is $\phi_i$.
\end{definition}

Given an instance $\mathbf{x} = [\ldots, x_j, \ldots]$ in a dataset $D$, 
the key to calculating the influence of feature $x_j$ to other features in $\mathbf{x}$, in terms of their contributions to the prediction, is by considering a hypothetical instance $\mathbf{x}' = [\ldots, x_j', \ldots]$ in the dataset $D^{-j}$ where $x_j$ is replaced by $x_j'$, the mean of feature $j$ in $D$. Replacing all feature $j$ values with their means in the dataset effectively "nullifies" the contribution of feature $j$ to the prediction
in all instances. Formally, we define the notion of {\em reduced dataset} as follows.



\begin{definition}
  \label{dfn:reduced_dataset}

  Given a dataset $D = \{([x_{11}, \ldots, x_{1m}], y_1), \ldots,
  ([x_{n1}, \ldots,x_{nm}], y_n)\}$,  a {\em
    reduced dataset} of $D$ with respect to feature $j (1 \leq j \leq m)$
  is $D^{-j} = \{([x_{11}^{-j}, \ldots, \\ x_{1m}^{-j}], y_1), \ldots, 
  ([x_{n1}^{-j}, \ldots, x_{nm}^{-j}], y_n)\}$ where
  \begin{equation*}
    x_{ik}^{-j} =
    \begin{cases}
      x_{ik}, & \mbox{if } k \neq j \mbox{;}\\
      \frac{1}{n}\sum_{1\leq i \leq n} x_{ik}, & \mbox{otherwise}.
    \end{cases}
  \end{equation*}
\end{definition}

With a reduced dataset defined, we can now define {\em peer-influence explanation} describing the influence of one feature to the others as follows.

\begin{definition}
  \label{dfn:peer-influence-exp}
Given an instance $\mathbf{x} = [x_1, \ldots, x_m]$ from some dataset $D$, 
for a prediction model $f$ and a feature attribution explainer $g$,
let $\phi_i$ and $\phi_i^{-j} (1 \leq i \leq m)$ be the feature
explanations of $x_i$ computed using $g$ with respect to $D$ and
$D^{-j}$, respectively. A {\em peer-influence explanation (PI-Explanation)}
${E}_{\mathbf{x}}$ (of $f(\mathbf{x})$ computed using $g$) is an
$m$-by-$m$ matrix where each element ${E}_{\mathbf{x}}[i,j] = \phi_i - \phi_i^j 
$. 
\end{definition}


From Definition~\ref{dfn:peer-influence-exp}, we see that a PI-Explanation is a square matrix. This matrix is asymmetric due to the nature of interaction, inferring the difference in Shapley values are non commutative across the matrix. 

To illustrate PI-Explanations, we consider an example drawn from the Simulacrum data set\footnote[1]{\url{https://simulacrum.healthdatainsight.org.uk/}}, developed by anonymously replicating the National Cancer Registration and Analysis Service data set. For illustration, we focus on lung cancer patients as indicated by their ICD-10 code ``C34'' (malignant neoplasm of bronchus and lung), and obtain a data sample with 2493 patients. Each patient is characterised by five features: ``M Best'' (presence or absence of distant metastatic spread), ``N Best'' (extent of involvement of regional lymph nodes), ``Weight'', ``Age'' and ``Height''. A binary classification, whether the life expectancy of the patent is greater than six months, is considered. An XGBoost classifier is used as the prediction model and on a 70/30 training/testing split, we achieve a classification accuracy 95.72\%. 

\begin{table}[ht] 
\caption{A lung cancer patient instance drawn from the Simulacrum data set.\label{table:patientInstance}}
\centering
\begin{scriptsize}
\begin{tabular}{|c|c|c|c|} 
 \hline
  \textbf{Patient Feature} & \textbf{Feature Value} & \textbf{Feature Type} & \textbf{Feature Attribution} \\ 
 \hline
 \hline
    M Best      & 1b   & Categorical & 3.24\\    
    N Best      & 0    & Categorical & -0.01\\
    Weight (kg) & 67   & Numerical (Controllable) & -0.65\\
    Age         & 65   & Numerical & 1.71\\
    Height (m)  & 1.78 & Numerical & 1.88\\    
 \hline
 \hline
 \multicolumn{4}{|l|}{\textbf{Prediction:} 0} \\
 \hline
\end{tabular}
\end{scriptsize}
\end{table}

Given the patient shown in Table~\ref{table:patientInstance}, its prediction outcome is 0, indicating that the predicted life expectancy of the patient is less than 6 months. The feature attribution explanation calculated using SHAP is 
shown in the right-most column.
The PI-Explanation $E_\mathbf{x}$ of this instance is shown below. Note that PI-Explanation differs from the correlation matrix of features ($M$ shown below). $E_\mathbf{x}$ is ``local'' to the given instance $\mathbf{x}$ for its prediction, whereas $M$ is a ``global'' measure over the entire data set.

\begin{scriptsize}
$$ E_{\mathbf{x}} = \begin{pmatrix}
0 & -2.92 & 2.38 & -0.88 & 4.18  \\ 
1.68 & 0 & 1.85 & -1.30 & 2.97  \\ 
2.31 & -2.25 & 0 & -2.04 & 3.55		\\ 
1.82 & -2.60 & 1.41 & 0 & 3.66\\ 
1.78 & -2.08 & 0.66 & -0.77	& 0 \\ 
\end{pmatrix}
\qquad
M = \begin{pmatrix}
1 & 0.004 & -0.06 & 0.09 & 0.07 \\ 
0.004 & 1 & -0.02 & 0.03 & 0.06 \\ 
-0.06 & -0.02 & 1 & 0.08 & -0.02 \\ 
0.09 & 0.03 & 0.08 & 1 & 0.26 \\ 
0.07 & 0.06 & -0.02 & 0.26 & 1 \\ 
\end{pmatrix}
$$
\end{scriptsize}

Intuitively, a positive $E_\mathbf{x}[i,j]$ means that the presence of a feature $i$, taking its current value of $x_i$ supports the feature $j$ with its value $x_j$ in predicting $f(\mathbf{x})$; a negative $E_\mathbf{x}[i,j]$ infers the opposite. For an instance, its PI-Explanation paints a more comprehensive picture of a features influence towards the prediction, with respect to the contribution of other features. To better visualise such contribution, we propose a graphical representation, PI-Explanation graph $G_{\mathbf{x}}$.
%
%
For an instance $\mathbf{x}$ with $m$ features, $G_{\mathbf{x}}$ is a directed graph with a set of vertices $V$ containing $m$ feature labels, and a set of arcs $A$ over $V$. The PI-Explanation $E_\mathbf{x}$ defines the relationship between vertices in $V$. More formally,

\begin{definition}
  \label{dfn:peer-influence-graph}

  Given a PI-Explanation ${E}_{\mathbf{x}}$ for an
  instance $\mathbf{x}$ with feature attribution explanations $\phi =
  [\phi_1,\ldots,\phi_m]$, let $F = [f_1, \ldots, f_m]$ be the list of feature labels, then the {\em Peer-Influence Graph (PI-Graph)} of
  ${E}_{\mathbf{x}}$ is a directed graph ${G}_{\mathbf{x}} = \langle
  V, A \rangle$  such that

  \begin{itemize}
  \item
    $V = V^p \cup V^o$ is the set of vertices where
    \begin{itemize}
    \item
      $V^p = \{f_i \in F|i \in \{1, \ldots, m\}, \phi_i \geq 0\}$, and
    \item
      $V^o = \{f_i \in F|i \in \{1, \ldots, m\}, \phi_i < 0\}$;
    \end{itemize}

  \item
    $A = A^+ \cup A^-$ is the set of arcs where
    \begin{itemize}
    \item
      $A^+ = \{(v_i,v_j)|v_i, v_j \in V, {E}_{\mathbf{x}}[i,j] \geq 0\}$, and
    \item
      $A^- = \{(v_i,v_j)|v_i, v_j \in V, {E}_{\mathbf{x}}[i,j] < 0 \}$.
    \end{itemize}
  \end{itemize}
  $V^p$ and $V^o$ are referred to as proponent and 
  opponent vertices, respectively; and features labelling $V^p$ and $V^o$ are proponent and opponent features, respectively. 
  $A^+$ and $A^-$ are referred to as support and attack arcs, respectively. 
\end{definition}

\begin{figure}[ht]
\centering
\includegraphics[scale=0.35]{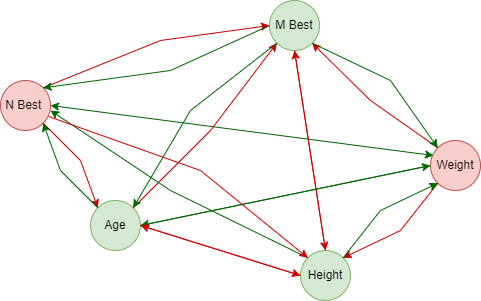}
\caption{The PI-Graph $G_\mathbf{x}$ of the instance $\mathbf{x}$ in the running example. Here, we see that the red vertices $V^o$ = \{N Best and Weight\} are the opponents, whereas the green vertices $V^p$ = \{M Best, Age, Height\} are the proponents. The green arcs are support arcs $A^-$; and red arcs are attack arcs $A^+$.}
\label{fig:peer-influnce-graph}
\end{figure}

Intuitively, features labelling $V^p$ support the prediction; whereas features labelling $V^o$ do not. An arc $(v_i, v_j)$ in 
$A^+$ indicates that the feature $v_i$ supports $v_j$ in making the prediction. Thus, arcs represent a ``second order effect'' of features in making predictions, through supporting or attacking other features. Continuing the running example, the PI-Graph $G_{\mathbf{x}}$ is shown in Figure~\ref{fig:peer-influnce-graph}.

From PI-Explanations and PI-Graphs, we can compute {\em alteration features} that recommend the most effective intervention to a prediction. In the context of the running example, we can ask: {\em which features, if changed, would be the most effective in altering the outcome of the prediction - letting the patient to survive for more than 6 months.} To this end, we want to find the feature that is most supportive to opponent vertices and offensive to proponent vertices. Formally,

\begin{definition}
\label{dfn:alterationIndex}

Given an instance $\mathbf{x}$ with $m$ features, PI-Explanation $E_\mathbf{x}$, and PI-Graph $G_\mathbf{x}$, let $P$ and $O$ be the sets of indices of proponent and opponent features, respectively. The {\em alteration index (ALT)} is defined, such that

\begin{equation}
    ALT_\mathbf{x} = \underset{i \in \{1,\ldots,m\}}{\argmin} \left(\sum_{i \in V} E_{\mathbf{x}}[i,j] \right).
\end{equation}
\end{definition}

With our running example, the process of computing $ALT_\mathbf{x}$ is shown in Table~\ref{table:alt}. We identify that changing 
feature $Age$ is most effective in altering the prediction outcome - tending the patient towards surviving for more than 6 months.

\begin{table}[]
    \centering
    \begin{tabular}{|c|c|c|c|}
    \hline
        $i$ & Feature Name & Proponent / Opponent & $\sum_{i \in V} E_{\mathbf{x}}[i,j]$ \\
    \hline
    \hline
         1 & Age  & $P$ &  0 + -2.92 + 2.38 + -0.88 + 4.18 = 2.76\\
         2 & Weight & $O$ &  1.68 + 0 + 1.85 + -1.3 + 2.97 = 5.2\\
         3 & Height & $P$ &  2.31 + -2.25 + 0 + -2.04 + 3.55 = 1.57 \\
         4 & N Best & $O$ &  1.82 + -2.6 + 1.41 + 0 + 3.66 = 4.29\\
         5 & M Best & $P$ &  1.78 + -2.08 + 0.66 + -0.77 + 0 = -0.41\\
    \hline
    \end{tabular}
    \caption{$ALT_{\mathbf{x}}$ calculation with all features. 
    From this table, we see that the features $M Best$ is the most effective feature in altering the outcome of the prediction}
    \label{table:alt}
\end{table}

From the PI-Explanation we can observe feature interactions for a given prediction. To identify the most suitable feature to alter for reaching opposing class and to suggest suitable interventions, we introduce the concept of a conflict matrix $C_x$ to avoid potential skew from attribution values and instead, focus on the relationships provided by arcs between the vertices.

\begin{definition}
  \label{dfn:conflictMatrix}
 
  Given an PI-Explanation $E_\mathbf{x}$, and alteration index $ALT_x$, we define a conflict matrix $C_\mathbf{x}$ as an $m$-by-$m$ matrix, such that \\
  \begin{equation*}
  C_\mathbf{x}[i,j] =
  \begin{cases}
  1, & \mbox{if } E_\mathbf{x}[i,j] > 0; \\ 
  -1, & \mbox{otherwise}.
  \end{cases}
  \end{equation*}
\end{definition}

 This provides either a single or set of features that can then through a conflict alteration index (CALT) that can identify the features that be altered to reach the opposing predicted class, introduced as follows. 
 
\begin{definition}
Given an instance $\mathbf{x}$ with $m$ features, PI-Explanation $E_\mathbf{x}$, alteration index $ALT_x$, and conflict matrix $C_x$. Let $P$ and $O$ be the sets of indices of proponent and opponent features, respectively. The {\em conflict alteration index (CALT)} is defined, such that
\begin{equation}
    CALT_\mathbf{x} = \underset{i \in \{1,\ldots,m\}}{\argmin} \left(\sum_{i\in V} C_{\mathbf{x}}[i,j]\right).
\end{equation}
\end{definition}

\begin{table}[]
    \centering
    \begin{tabular}{|c|c|c|c|}
    \hline
        $i$ & Feature Name & Proponent / Opponent & $\sum_{i \in V} C_{\mathbf{x}}[i,j]$ \\
    \hline
     \hline
         1 & Age & $P$ &  1\\
         2 & Weight & $O$ & 3\\
         3 & Height & $P$ &  1\\
         4 & N Best & $O$ &  3\\
         5 & M Best & $P$ &  1\\
    \hline
    \end{tabular}
    \caption{$CALT_{\mathbf{x}}$ calculation with all features. 
    From this table, we see that the features $Age$, $Height$ and $M Best$ are the most effective features in altering the outcome of the prediction}
    \label{table:alt}
\end{table}

\section{Results}
\label{sec:result}

\subsection{Case Study (Confirming Alteration Index)}

We construct a data-set containing two more features ``T Best" (the size and extent of the main tumour) and ``Dose Administration", to demonstrate the method by introducing more permutable options from a medical perspective. An XGBoost classifier is used as the prediction model on a 70/30 training/test split, achieving a classification accuracy 96.59\%.

We introduce a second case, we explore explanations based on a ground truth. We view a patient instance in two given drug cycles to see if explanations can help drive decisions and whether or not this conformed to the actual intervention. For this case, we take a patient where in an early drug cycle is predicted the survival time of less than six months, but the ground truth is survival greater than six months at a latter cycle date.

We introduce both drug regimen administration cycles for a patient instance in Table \ref{table:patientInstance2}. Following this, we produce a feature attribution local explanation for cycle 2 of the patient instance, where we can observe prominence of the features towards each class, where we identify $Weight$ as the most important feature in the current prediction. We see that in cycle 3, medical intervention alters the variable $Dose Administration$, which as identified by feature attribution in both cycles holds little impact on the model. To gauge a deeper understanding of relationships we construct the framework to provide a CALT. 

\begin{table}[ht]
\caption{Patient Instance 2}
\resizebox{\textwidth}{!}{
\label{table:patientInstance2}
\centering
\begin{tabular}{|c|c|c|c|c|} 
 \hline
  \textbf{Patient Feat.}  & \textbf{Feat. Value (Cycle 2)} & \textbf{Feat. Attribution} & \textbf{Feat. Value (Cycle 3)} & \textbf{Feat. Attribution} \\ 
 \hline\hline
    Dose Administration &  1000 & -0.09 &  90 & -0.01 \\ 
    M Best &  1 &  1.52 &  1 & 2.23\\    
    N Best &  2 & 1.10  &  2 & 0.05\\ 
    T Best &  4 & -0.14 &   4 & -0.42\\ 
    Weight &  76 & 3.09 &  76 & -3.02\\ 
    Age &  69 & -0.01 &  69 & -1.38\\ 
    Height & 1.75m & 0.63 & 1.75m & -0.10\\    
 \hline\hline
 \multicolumn{5}{|l|}{\textbf{Prediction Cycle 2:} 0} \\ 
 \multicolumn{5}{|l|}{\textbf{Prediction Cycle 3:} 1} \\
 \hline
\end{tabular}
}
\end{table}


To provide a deeper insight of the relationships, we produce a PI-Explanation $E$, such that, 

$$ E_x = \begin{pmatrix}
0 & 2.01 & 1.03 & -0.08 & -0.86 & -0.65 & 1.52\\ 
-0.53 & 0 & 0.70 & -0.38 & -1.37 & -0.44 & 1.54 \\ 
-0.24 & -0.51 & 0 & -0.27 & -0.64 & 0.26 & 1.54 \\ 
-0.07 & -0.23 & -0.37 & 0 & -0.89 & 0.31 & 1.53\\ 
0.24 & -0.16 & -0.73 & -0.30 & 0 & 0.06 & 1.52  \\ 
0.08 & -0.52 & 0.46 & 0.19 & -0.39 & 0 & 1.53 \\ 
-0.80 & -0.25 & 0.36 & -0.03 & -0.42 & 0.03 & 0 \\ 
\end{pmatrix}$$

\begin{table}[]
    \centering
    \begin{tabular}{|c|c|c|c|}
    \hline
        $i$ & Feature Name & Proponent / Opponent & $\sum_{i\in V} E_{\mathbf{x}}[i,j]$ \\
    \hline
    \hline
         
         1 & Age & $O$ &  2.97 \\
         2 & Weight & $P$ & -0.48\\
         3 & Height & $P$ & 0.14 \\
         4 & Dose Administration & $O$ & 0.28\\
         5 & T Best & $O$ &  0.63\\
         6 & N Best & $P$ & 1.35\\
         7 & M Best & $P$ &  -1.11 \\
    \hline
    \end{tabular}
    \caption{$ALT_{\mathbf{x}}$ calculation with all features. 
    From this table, we see that the features $M Best$ is the most effective feature in altering the outcome of the prediction.}
    \label{table:alt2}
\end{table}

Posterior to this, we produce the conflict matrix $C_x$ to map to the CALT to indicate which features from cycle 2 can be altered to reach the opposing prediction. The resulting set as illustrated in \ref{table:calt2} is to change a feature from the set \{Weight, Height, Dose Administration, M Best\}. To analyse the given result, we view patient cycle 3 to see the medical intervention that changed the outcome and the prediction to be true for $>$ 6 Months survival.

In the true intervention, the medical professional administered change in dose administration, which greatly shifted the prediction of the model, a feature also identified by relationships provided by the conflict matrix. 

\begin{table}[]
    \centering
    \begin{tabular}{|c|c|c|c|}
    \hline
        $i$ & Feature Name & Proponent / Opponent & $\sum_{i\in V} C_{\mathbf{x}}[i,j]$ \\
    \hline
    \hline
          1 & Age & $O$ &   -1 \\
          2 & Weight & $P$ &  -3\\
          3 & Height & $P$ &  -3 \\
          4 & Dose Administration & $O$ & -3\\
          5 & T Best & $O$ &  -1\\
          6 & N Best & $P$ &   1\\
          7 & M Best & $P$ &  -3\\
    
    \hline
    \end{tabular}
    \caption{$CALT_{\mathbf{x}}$ calculation with all features. 
    From this table, we see that the feature set $M Best$, $Dose Administration$, $Weight$ and $Height$ are the best feature selection that can be altered to change prediction based on the  relationships between vertices.}
    \label{table:calt2}
\end{table}



\begin{figure}[ht]
\centering
\includegraphics[scale=0.3]{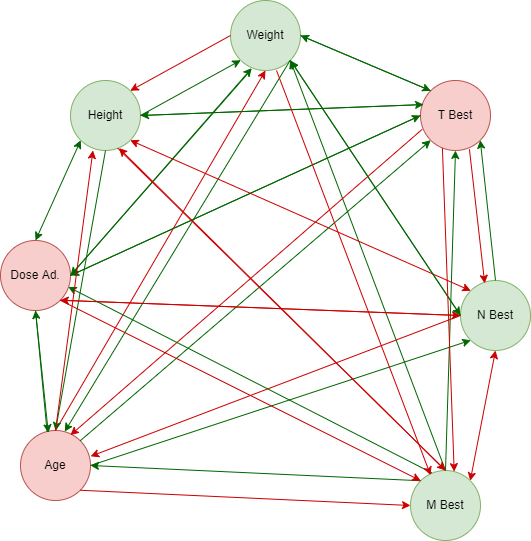}
\caption{A PI-Graph $G_\mathbf{x}$ provided for patient instance 2 - in drug cycle 2.}
\label{Graph3}
\end{figure}

The purpose behind interjecting meaning behind a relationship during an earlier cycle is to ensure the CALT adheres to domain knowledge. As such, we construct the PI-Graph $G_\mathbf{x}$ which should draw insight towards the relationship of the latter cycle 3. Therefore, from the CALT we determine our relationships between vertices and the associated arcs to map the graph in Figure \ref{Graph3}.



Observing the patient instance, we provide PI-graph for a local explanation. The PI-Explanation provides a form of reasoning as to why feature attribution outcomes can drastically changed irrespective of the fact ``Dose Administration" according to feature attribution has little to no impact. We determine that, although this has no independent impact as identified by feature attribution methods, the impact on $Dose Administration$ peers represented by relationships in the graph and the latter conflict matrix, $Dose Administration$ is an effective alteration, indirectly holding great importance based off both the CALT and expert opinion.

\section{Related Work} 
\label{sec:related}

In \cite{Dejl}, the authors explore the notion of providing explanation from an argumentation framework as an extension of influence namely \emph{Deep Argumentative Explanations} (DAX), comparatively, our work introduces the idea behind attack and support relations but conversely focuses on the interjection of feature attribution values and the relationship introduces peer-influence explanations that focus on the impact each feature has on the remaining. \cite{SHAPFlow} introduce a Shapley Value based casual graph representation namely \emph{Shapley Flow} to understand the casual inference, as such, the work provides a flow of feature impact towards the model outcome and the associated Shapley Values, Compared to this our work instead is a tertiary extension that can be modelled for any feature attribution or feature importance method e.g. LIME, Native XGBoost Importance etc.  that doesn't rely on already having access to a complete causal graph.

Similarly, we see how argumentation can be used in a Bayesian setting \cite{Rago2021InfluenceDrivenEF} introducing \emph{Influence-Driven Explanations} displaying a form of Shapley Value influence on top of others using argumentation. Compared both of these, our work provides the bi-directional relationship of feature impact for a complete graph with a differing method of influence calculation and the exaggeration of peer-influence explanations with the extension of defining an alteration index for reaching a desired outcome.

The work \cite{Sarker} introduces graph based explanations through knowledge graph generation using ontological knowledge. This presents a hierarchical structure that can be used to understand relationships between each feature. Unlike this work, the explanations that we provide are focused on the feature importance values returned through  the black-box model itself or the feature attribution values provided by model-agnostic methods, in the case of this work we use Shapley Values. In \cite{GBEx} we see the introduction of Graph Based Explanations namely GBEx, which presents a form of generalised importance between features where edges represent the importance between each of the features. Conversely, in our work we demonstrate the direction to which the feature impacts the next with an associative attribution value which can conditionally be defined as attack or support relationships with the extension of alteration index identification.

\section{Conclusion}
\label{sec:cln}

From the current implementation of XAI methods, it's required that tertiary human domain knowledge is used in order to leverage the use of explanations in the medical field. Ideally, explanations should form a source reasoning between the underlying features and aim to provide granular reasoning to an outcome much like we can as humans. As such, the explanation should draw from variable influence to determine a complete explanation across all features in a model. 

Given that the results conform to domain knowledge, we make the assumption that the baseline implementation and further development of this work can provide a framework that will provide knowledge of unseen features and the underlying connection between features, with graphical support, as well as existing features. The method introduced in this paper provides a form of peer-influence, paving the way to further development through optimisation and visualisation. 

Conclusively, the existing framework provides a baseline to view peer-influence and inform perturbations through the constructed relationships and the modelling of the peer-influence graph $G_\mathbf{x}$. We likewise reflect on Argumentation Frameworks \cite{DUNG1995321} as a potential avenue for extension, with both a Bipolar Argumentation Framework \cite{Cayrol2009} and (Audience) Value-Based Argumentation Framework \cite{BenchCapon}\cite{BenchCapon2} holding merit towards future work as we observe properties that may allow for the adoption of Argumentation Frameworks. Similarly, we aim to proceed with optimisation to the perturbations in future work, as such, we introduce the preliminary framework of this paper behind intution and what can be seen through methodic intervention of explainable models.

\bibliographystyle{unsrt} 

\bibliography{GRAPHShap}

\end{document}